\documentclass[10pt,twocolumn,letterpaper]{article}

\usepackage{iccv}
\usepackage{times}
\usepackage{epsfig}
\usepackage{graphicx}
\usepackage{subfigure}
\usepackage{amsmath}
\usepackage{amssymb}
\usepackage{booktabs}
\usepackage{algorithm}
\usepackage{algorithmic}
\usepackage{multirow}
\usepackage{color}
\usepackage{booktabs}
\usepackage{enumerate}
\usepackage[small]{caption}
\definecolor{aa}{rgb}{1,0,0}
\definecolor{bb}{rgb}{0,0,1}

\usepackage[breaklinks=true,bookmarks=false]{hyperref}

\iccvfinalcopy % *** Uncomment this line for the final submission

\def\iccvPaperID{****} % *** Enter the ICCV Paper ID here

\begin{document}

%%%%%%%%% TITLE
\title{Deep Reasoning with Multi-Scale Context for Salient Object Detection}
\author{\normalsize{Zun~Li$^{1}$,  Congyan~Lang$^{1}$,
Yunpeng~Chen$^{2}$,
Jun Hao~Liew$^{2}$,
Jiashi~Feng$^{2}$} \\
	\small{$^{1}$Beijing Jiaotong University, $^{2}$National University of Singapore} \\
	{\small lznus2018@gmail.com,  cylang@bjtu.edu.cn,
	chenyunpeng@u.nus.edu,%chenyunpeng@u.nus.edu,
	{\small liewjunhao@u.nus.edu,
	elefjia@nus.edu.sg}}
	}
\maketitle
%\thispagestyle{empty}

%%%%%%%%% ABSTRACT
\begin{abstract}
To detect salient objects accurately,
existing methods usually design complex backbone network architectures
to learn and fuse powerful features.
However, the saliency inference module that performs saliency prediction from the fused features receives much less
attention on its architecture design and typically adopts only a few fully convolutional layers.
In this paper,
we find the limited capacity  of the saliency inference module indeed makes a fundamental performance bottleneck, and enhancing its capacity is critical for obtaining better saliency prediction. Correspondingly, we propose a deep yet light-weight saliency inference module that adopts a multi-dilated depth-wise convolution architecture. Such a deep inference module, though with simple architecture, can directly perform reasoning about salient objects from the  multi-scale convolutional features fast, and give superior salient object detection performance with less computational cost.
To our best knowledge, we are the first to reveal the importance of the inference module for salient object detection, and present a novel architecture design with attractive efficiency and accuracy.  Extensive experimental evaluations demonstrate that our simple framework
performs favorably compared with the state-of-the-art methods with complex backbone design.
\end{abstract}

%%%%%%%%% BODY TEX
\section{Introduction}
Salient object detection aims to identify the most visually conspicuous objects in an image, and is an important pre-processing step for various computer vision applications, such as image segmentation~\cite{Wei2015STC},
image understanding~\cite{understanding}, image captioning~\cite{captionI,captionII} and visual tracking~\cite{tracking}.
Early methods~\cite{DUT-OMRYang,DRFI,GC,RBD} generally utilize hand-crafted visual features and heuristic clues, which have limited capacity of modeling and describing high-level semantics.
Recently,
convolutional neural networks (CNNs),  especially the fully convolutional networks (FCNs),
have been extensively utilized  to learn more powerful features for salient object detection.
\begin{figure}[t]
\begin{center}
%{\includegraphics[width=0.48\textwidth]{pictures/introduce2.png}}
{\includegraphics[width=0.47\textwidth]{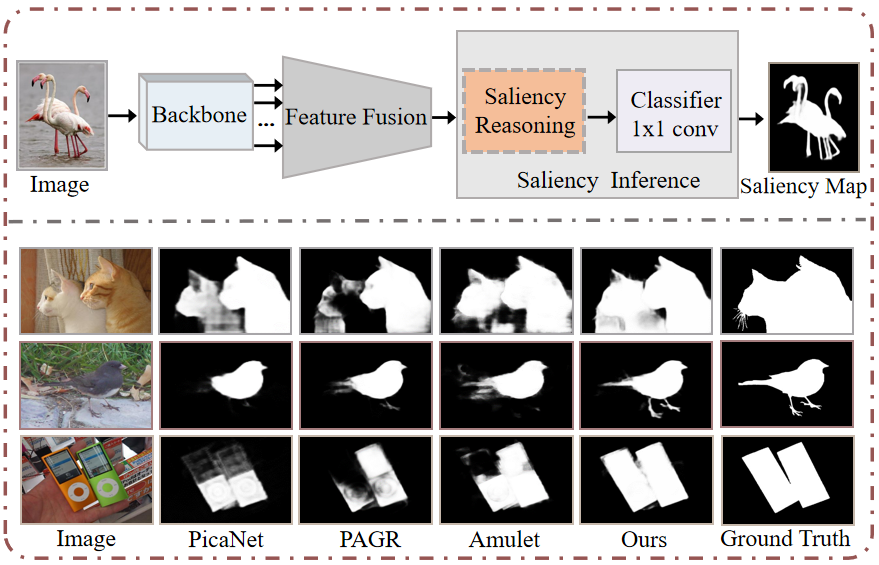}}
\end{center}
\vspace{-5mm}
\caption{
Top panel: existing FCN-based saliency detection pipeline. It usually includes three components: backone network, feature fusion and saliency inference. Bottom Panel: examples of saliency maps produced by PicaNet~\cite{PiCANet:liu2018picanet},
PAGR~\cite{PAGR:zhang2018progressive}, Amulet~\cite{Amulet:zhang2017amulet} and ours.
In this work, we focus on the \textit{Saliency Reasoning} part which is the core of the saliency inference component.
Our method
is consistently better than the state-of-the-art methods.}
\label{intro}
\vspace{-4mm}
\end{figure}

Existing FCN-based models typically consist of three components: the backbone, feature fusion and saliency inference, as shown in the top panel of Fig.~\ref{intro}.
Given an image, the backbone network produces a set of feature maps with different spatial scales, and the
feature fusion component integrates these multi-scale features to form  a discriminative image
representation. Such a representation is then fed into a saliency inference component that performs saliency reasoning and a classifier with $1\times1$ convolution to generate the saliency detection result.

Among the three components,
the feature fusion receives much attention in recent
FCN-based saliency detection works~\cite{DGRL:wang2018detect,PiCANet:liu2018picanet,Amulet:zhang2017amulet,BMPM:zhang2018bi,
SRM:wang2017stagewise,ELD:lee2016deep,NLDF:luo2017non,RFCN:wang2016saliency,RAS:chen2018eccv,R3Net:deng18r,PAGR:zhang2018progressive}.
These works design complex and powerful architectures to fuse
the multi-scale features, extensively leveraging intermediate supervision for each FCN layer.
For example,
in~\cite{PiCANet:liu2018picanet},
two PicaNets are embedded into U-Net architecture to
incorporate global and local contexts at each decoding module,
and the final saliency is predicted with weighted deep supervision on these modules.
These previous methods have achieved impressive results.
However, existing fusion-heavy models cannot well handle the various challenges presented in the images,
as shown in Fig.~\ref{intro}.
Such as objects touching image boundaries (1st row of bottom panel),
objects with similar appearance to background (2nd row of bottom panel),
and images with complex background and foreground (3rd row of bottom).

\begin{figure*}[t]
\begin{center}
%{\includegraphics[width=0.95\textwidth]{pictures/frame4.png}}
{\includegraphics[width=0.95\textwidth]{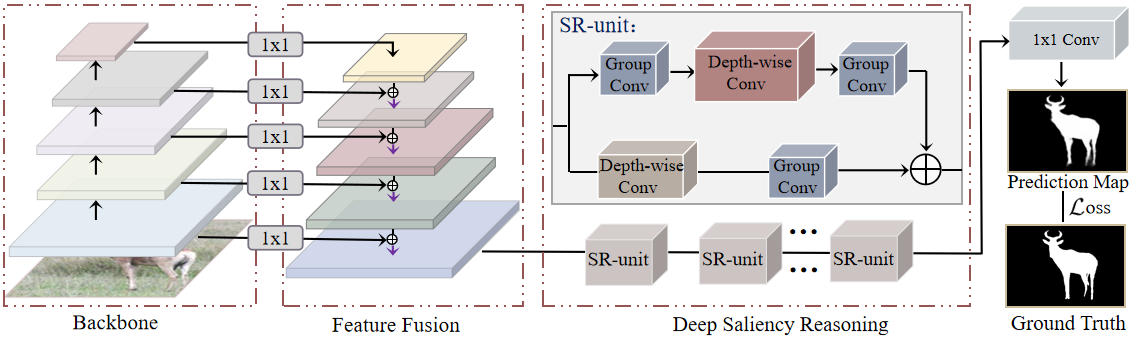}}
\end{center}
\vspace{-5mm}
\caption{Saliency detection framework with our proposed saliency reasoning module.
This framework first fuses comprehensively saliency features of backbone feature extraction networks
with upsampling and concatenation,
and then directly reasons over these features via designing a deep multi-dilated depth-wise convolution architecture (see Fig.~\ref{reasoningExample}).}\label{framework}
\end{figure*}

We find the performance bottleneck actually lies in the salieny inference component which however receives much less attention before.
For this component,
existing methods only use a few fully convolutional layers to perform the saliency reasoning,
without effective information communication between multi-scale features.
Accordingly,
we propose to further boost saliency detection performance by enhancing capacity of the inference.
We consider that the backbone network in Fig.~\ref{intro}
already offers comprehensive multi-scale saliency features, and more effort should be devoted to developing proper architectures for the inference part to maximally reason about
saliency prediction from these features.
Some recent works have demonstrated the importance of model reasoning ability through adopting graph convolution~\cite{GlorNet}\cite{wang2018videos}, incorporating higher-order information~\cite{lin2015bilinear}, \emph{etc}.
However, they are typically complicated and cannot be directly applied to
the saliency detection task.

In this work, we develop a simple yet effective inference architecture  with an enhanced reasoning ability for salient object detection, by stacking multiple dilated-convolution layers to form a deep saliency reasoning module.
To reduce computation overhead
and improve salient object detection in complex scenes,
the reasoning module fully exploits the light-weight pointwise group convolutions and depth-wise convolutions, thus costs less time
while offering superior saliency prediction
accuracy. The overall architecture of our proposed model is shown in Fig.~\ref{framework}. It is compatible with  any popular backbone network, such as ResNet~\cite{resnet:He2015Deep}, VGG~\cite{VGG},
to extract multi-scale saliency features.
Then it directly fuses these features via a top-down pathway and lateral connections.
Afterwards,
it performs deep saliency reasoning
by multi-dilated depth-wise convolution units
over the fused features and predicts the saliency maps via simple $1\times 1$ convolution. The main contributions of this work are summarized as follows:

\begin{itemize}
\setlength\itemsep{0em}
\item To our best knowledge, we are the first to uncover the importance  of the saliency inference component which has been neglected by most saliency detection models.
\item We propose a simple yet effective deep reasoning module to better infer  saliency predictions from multi-scale saliency features,
with less computational cost and superior performance.
\item We conduct comprehensive experiments to compare the network with our proposed reasoning module and recent state-of-the-art methods.
Our network outperforms well established baselines under various metrics significantly.
\end{itemize}

\section{Related work}
\label{relatedwork}
Early saliency detection methods usually extract hand-crafted visual features
(\emph{e.g.}, color~\cite{PSACALSLi}, texture~\cite{DUT-OMRYang}, intensity contrast~\cite{DRFI}),
and then classify them into salient and non-salient ones.
Some heuristic saliency priors are also utilized including color contrast~\cite{FT,GC}, center prior~\cite{DRFI,DSR}
and background prior~\cite{RBD,DUT-OMRYang,GS}.
Recently, deep CNNs have been extensively employed for saliency detection due to their
strong representation learning capability.
For instance,
Wang \etal~\cite{LEGS:wang2015deep} proposed two CNNs to
aggregate local patch estimation and global proposal search to detect salient objects.
Li \etal~\cite{MDF:li2015visual} extracted multi-scale features and predicted  saliency for each image segment by  a fully-connected regressor network.
Zhao \etal~\cite{MCDL:zhao2015saliency} introduced the multi-context CNNs that exploit both local and global context for saliency prediction per superpixel.
Though with better performance than early methods,
these CNN-based models predict saliency at patch level, suffering severe artifacts and high computational cost.

%FCN-based saliency methods;
Most recent works~\cite{DS:li2016deepsaliency,DHSNet:liu2016dhsnet,Amulet:zhang2017amulet,RFCN:wang2016saliency,SRM:wang2017stagewise,PAGR:zhang2018progressive,DSS:hou2017deeply,UCF:zhang2017learning,PiCANet:liu2018picanet,NLDF:luo2017non,DGRL:wang2018detect,RAS:chen2018eccv,SDF} build models based on fully convolutional networks (FCNs) that make saliency prediction over the whole image directly.
For example,
Li \etal~\cite{DS:li2016deepsaliency} proposed a multi-task FCN for saliency detection.
Liu \etal~\cite{DHSNet:liu2016dhsnet} presented a deep hierarchical saliency network to learn
global structures and progressively refine the saliency maps via integrating local context information.
More recently,
Wang \etal~\cite{SRM:wang2017stagewise} proposed to
generate a coarse prediction map via FCN,
and then refine it stage-wisely.
Zhang \etal~\cite{PAGR:zhang2018progressive} introduced an attention guided network
that progressively integrates multiple layer-wise attention
for saliency detection.
Different from these methods that aggregate multi-level features stage-wisely,
some other works integrate multi-level features simultaneously.
Zhang \etal~\cite{Amulet:zhang2017amulet}
proposed to simultaneously aggregate multi-level feature maps
and perform saliency detection via  a bidirectional inference.
Hou \etal~\cite{DSS:hou2017deeply} introduced short connections to the HED~\cite{HED} architecture, and predicted saliency based on aggregated saliency maps from each side-output.
Zhang \etal~\cite{BMPM:zhang2018bi}  designed a bi-directional architecture  to extract
multi-level features and combine them to predict saliency maps.

Most works focus on designing complex feature learning and fusion modules, and adopt very simple saliency inference modules over the fused features.
For example,~\cite{NLDF:luo2017non,Amulet:zhang2017amulet} adopt
two $1\times1$ convolutional layers to infer saliency over the local and global features.
Some recent works use one~\cite{DGRL:wang2018detect} or two~\cite{DSS:hou2017deeply,SRM:wang2017stagewise,R3Net:deng18r} $3\times3$ convolutional layers to infer saliency per side-output.
Similarly,
Islam \etal~\cite{RSD} designed three $3\times3$ and one $1\times1$ convolutional layers
to predict saliency maps at each  refinement stage. Recent state-of-the-art model~\cite{PiCANet:liu2018picanet}  predicts the final saliency with one $1\times1$ convolutional layer.

Different from previous works,
we focus on saliency reasoning from the fused saliency features in this work. In particular,
we introduce a new architecture that  stacks multiple dilated depth-wise convolution layers to build a deep
saliency reasoning module.

\section{Proposed method}

\subsection{Deep saliency reasoning}
\label{reasoningMethodOne}
As shown in Fig.~\ref{intro}, most FCN-based saliency methods predict the saliency map $\mathrm{F}_S$ from the extracted features $S$ (after fusion) via the following saliency inference module:
%when given the fused features $S$, the saliency map of an image can be predicted by a saliency inference module, which is formulated as
\begin{flalign}
\mathrm{F}_{S}=\varphi\Big(g(S)\Big)
\end{flalign}
where $g(S)$ is the module  to reason about salient objects from the fused features $S$,
and $\varphi(\cdot)$  produces the final saliency map given  the saliency reasoning result.
Previous FCN based saliency models usually perform saliency reasoning with a few standard convolutional layers.
Such shallow architectures are only effective for simple reasoning tasks and incapable of conducting complex ones.

To increase the capacity of the saliency reasoning module and boost the overall saliency prediction performance, we propose to build a
\emph{deeper} and \emph{wider} reasoning module.
Inspired by the recent light-weight architecture design~\cite{ShuffleNet,ShuffleNetV2}, we propose to fully utilize group convolution and depth-wise convolution to better infer  saliency predictions from multi-scale saliency features,
with less computational cost and superior performance, which are revisited as follows.

\begin{itemize}%
\setlength\itemsep{0em}
\item \emph{Group Convolution} \cite{AlexNet}\cite{SqueezeNet}\cite{resnext:Xie2016}\cite{ShuffleNet}  evenly slices the input and output feature tensors into groups channel-wisely. The connections between different groups are removed. This leads to a sparsely connected convolution layer, which helps reduce both the computational cost and over-fitting risks.
\item \emph{Depth-wise Convolution} \cite{Xception} \cite{MobileNets} \cite{MobileNetV2}\cite{ShuffleNetV2}  is a special case of group convolution, where the number of groups equals that of channels. It performs spatial convolution with each channel of an input tensor separately.
\end{itemize}
The above two convolution operations can reduce the computational cost significantly. Given a convolution layer with input/out channel dimension of $C$, the complexity of the regular convolution
layer, group convolution layer, and depth-wise convolution layer is    $\mathcal{O}(C^2)$,  $\mathcal{O}(C/\# groups)$,
and $\mathcal{O}(C)$ respectively. Thus, adopting these sparse convolutions  can help build a deeper inference module with a stronger reasoning ability yet only bringing negligible computational overhead.

\subsection{SRNet}
\label{instantedNet}
We enhance the reasoning ability of the detection model over salient regions by stacking multiple computationally
efficient depth-wise convolutional layers systematically.
We here apply our proposed saliency reasoning to a deep saliency reasoning network which is named SRNet.
Fig.~\ref{framework} shows its overall architecture.
\vspace{-2mm}
\begin{figure*}[!t]
\begin{center}
{\includegraphics[width=0.95\textwidth]{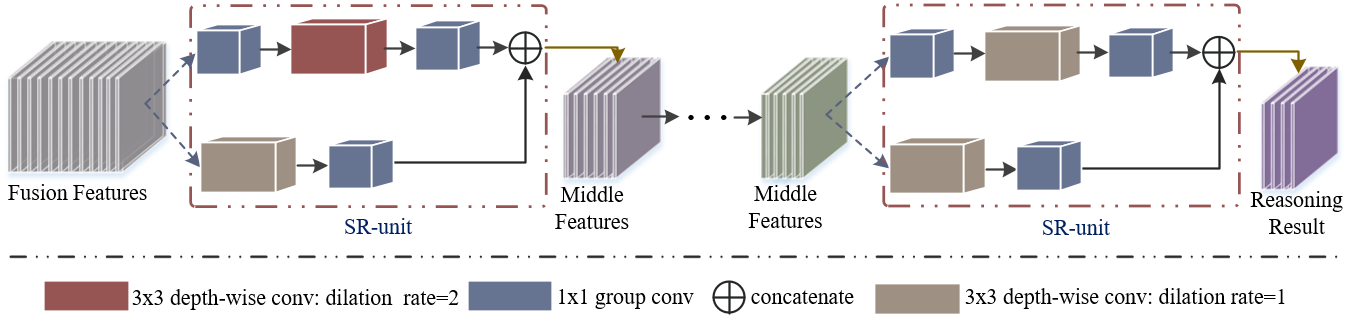}}
\end{center}
\vspace{-4mm}
\captionsetup{margin=10pt,justification=justified}
\caption{Illustration of saliency reasoning in SRNet. It takes multi-scale fused features as input, and performs saliency reasoning with several repeated SR-unit to reason about salient features. Each SR-unit contains some convolution and dilated depth-wise convolution operations.}\label{reasoningExample}
\end{figure*}
\vspace{-1mm}
\paragraph{Backbone network}
Our proposed saliency reasoning module is compatible with any popular backbone architectures. Here,
we adopt the  VGG16~\cite{VGG} and ResNet-101~\cite{resnet:He2015Deep}
as the backbones.
For VGG16,
in order to preserve relatively large spatial resolution in top layers,
following~\cite{PiCANet:liu2018picanet},
we append $1024$ $3\times3$ kernels with dilation of $8$ and $1024$ $1\times1$ kernels
to $conv5$ to replace the fully connected layers fc$6$ and fc$7$.
Then,
we utilize five convolutional blocks, including $\{conv2, conv3, conv4, conv5, conv7\}$, to extract
hierarchical saliency features.
For ResNet-101,
we remove its final fully-connected layer,
and utilize the first five convolutional blocks to extract multi-scale features.
For simplicity,
we uniformly denote the output feature tensors from the five blocks  as $\{S^{1},S^{2},S^{3},S^{4},S^{5}\}$ from
bottom to top,
with channels of $\{64,256,512,1024,2048\}$ in ResNet-101 and $\{128,256,512,512,1024\}$
in VGG16. The feature fusion introduced below is then performed on these hierarchical feature maps from $S^1$ to $S^5$.
\vspace{-2mm}
\paragraph{Hierarchical feature fusion}
For saliency detection,
high-level features help classify image regions while low-level ones help generate sharp and
accurate object boundaries.
To benefit from both   desired properties,
we introduce a fusion component to aggregate the multi-level saliency features
via a top-down pathway and lateral connections.

Concretely,
we first connect a $1\times1$ convolution layer after each convolutional block in the backbone,
and reduce their output feature channels to $\{64,128,256,256,256\}$
and $\{128,128,256,256,256\}$ in ResNet and VGG respectively.
Then, for any two adjacent blocks $\{S^{i}, S^{i+1}\}(i=1,2,3,4)$ in ResNet,
we first upsample $S^{i+1}$ by a factor of 2 via a bilinear interpolation operation,
and then concatenate this resulting feature map with $S^{i}$ directly, giving the fused feature map.
After this,
a $3\times 3$ convolution layer is applied to reduce the channels
of the fused feature map.
This feature fusion process is formulated as
\begin{flalign}
S^{i} &= \mathrm{conv}_{3\times 3}(S^{i}\oplus \mathrm{upsample}(S^{i+1})), \quad \forall i \in\{1,2,3,4\}. \nonumber
\end{flalign}
Similar operations are applied for $\{S^{1},S^{2},S^{3},S^{4},S^{5}\}$ in VGG, without the upsampling operation on $\{S^{3},S^{4},S^{5}\}$. Finally,
we take the output of $S^{1}$ with $128$ channels as the output fused saliency feature. The above fusion operation gradually integrates the low-level details into the high-level semantic-rich feature,
providing a high-quality image representation for accurate salient object detection.

\vspace{-2mm}
\paragraph{Saliency reasoning module}

We implement the saliency reasoning module based on a light-weight network architecture,
shuffleNet \cite{ShuffleNet}\cite{ShuffleNetV2},
which consists several repeated shuffle unit structures to predict saliency maps from the fused features. We name each unit structure as \emph{SR-unit},
and illustrate the reasoning module in SRNet-R with Fig.~\ref{reasoningExample}.
It contains several SR-unit to gradually reduce the feature channels and obtain more powerful saliency features for inferring saliency predictions.
In this module,
each SR-unit uses a shortcut scheme that performs saliency reasoning over the fused features with two branches.
For the first branch,
the module applies two $1 \times 1$ group convolutions and one $3 \times 3$ depth-wise convolution on the shortcut path.
For the second branch,
a $3 \times 3$ depth-wise convolution and one $1 \times 1$ group convolution are adopted to
transform the channel dimension to match the shortcut path.
To preserve spatial resolution of the fused features,
the stride of all the convolutions are set as $1$.
At last,
SR-unit concatenates outputs from those two transformed branches,
and conducts the channel shuffle \cite{ShuffleNet}  to communicate cross-channel information
  to improve prediction accuracy.

Although each SR-unit can reason effectively over the fused features,
due to their limited receptive field sizes,
some non-salient object regions will be wrongly detected.
To address this issue,
we propose to augment the saliency reasoning module by
introducing \emph{multi-dilated} depth-wise convolutions.
As shown in Fig.~\ref{reasoningExample},
we use dilation rate 2 and 1 for $3 \times 3$ depth-wise convolution
and alternately insert them into the repeated SR-unit.
such varying dilation rates enlarge receptive field of the convolution kernels.
In this way,
the discriminative features from the adjacent highlighted
regions can be transferred to the salient-related
regions that have not been discovered,
and more powerful contextual can be enhanced for accurate saliency detection.
Based on the proposed saliency reasoning module,
the final saliency map is predicted by applying a $1\times1$ (w/o non-linearization)
convolution and softmax classifier onto the saliency reasoning %result.

%%%%
\vspace{-4mm}
\paragraph{Training of SRNet}
Given the training dataset $\mathcal{X}=\{(X_{i},Y_{i})\}_{i=1}^{N}$ with $N$ training pairs,
where $X_{i}=\{\mathrm{x}_{k}^{i},k=1,...T\}$ is the input image,
and $Y_{i}=\{\mathrm{y}_{k}^{i},k=1,...T\}$ is the corresponding ground-truth map with $T$ pixels of $X_{i}$.
For the ground-truth $Y_{i}$,
we denote $\mathrm{y}_{k}^{i}=1$ as the salient pixel,
and $\mathrm{y}_{k}^{i}=0$ as the non-salient pixel. For SRNet,
we denote $\textbf{W}$ as the feature extraction and feature fusion parameters,
$\omega$ as the saliency reasoning parameters,
and $\theta$ as the classifier parameters for the final saliency prediction.
Then, the loss function $\mathcal{L}(\cdot)$ for training SRNet is expressed as
\begin{flalign}\label{eq:lossfunc}
\mathcal{L}(\textbf{W}, \omega, &\theta) = -\delta\sum_{k\in Y_{+}}\textrm{log}\ \textrm{Pr}(\mathrm{y}_{k}=1|X;\textbf{W},\omega,\theta) \nonumber \\
                             &-(1-\delta)\sum_{k\in Y_{-}}\textrm{log}\ \textrm{Pr}(\mathrm{y}_{k}=0|X;\textbf{W},\omega,\theta),
\end{flalign}
where $Y_{+}$ and $Y_{-}$ refer to the salient and non-salient label sets, respectively.
$\delta$ is the loss weight to balance the losses between salient and non-salient pixels.
$\textrm{Pr}(\mathrm{y}_{k}=0|X;\textbf{W},\omega,\theta)$ is the probability score that
measures how likely the pixel belongs to the salient region.
In this work,
we compute the confidence score with the following softmax classifier function:
\begin{flalign}\label{eq:gailv}
\textrm{Pr}(\mathrm{y}_{k}=1|X;\textbf{W},\omega,\theta)=\frac{e^{z_{1}}}{e^{z_{0}}+e^{z_{1}}},\\
\textrm{Pr}(\mathrm{y}_{k}=0|X;\textbf{W},\omega,\theta)=\frac{e^{z_{0}}}{e^{z_{0}}+e^{z_{1}}},
\end{flalign}
where $z_{0}$ and $z_{1}$ denote
the score of non-salient and salient label, respectively.
Since Eq.~(\ref{eq:lossfunc}) is continuously differentiable,
we adopt stochastic gradient descent (SGD) method to train our network,
and the optimal parameters can be obtained by
\begin{flalign}\label{eq:SDG}
(\textbf{W}^{*}, \omega^{*}, &\theta^{*})=\textrm{arg}\ \textrm{min}\ \mathcal{L}(\textbf{W}, \omega, \theta).
\end{flalign}
With SGD,
SRNet is trained by feeding the fixed-size input images into the network,
and it directly predicts the final saliency map without any post-processing.

%%%state-of-the-art saliency maps
\begin{figure*}[!pt]
% \setcaptionmargin{0.1in}
\begin{center}
\subfigure
{\includegraphics[width=1.0\textwidth]{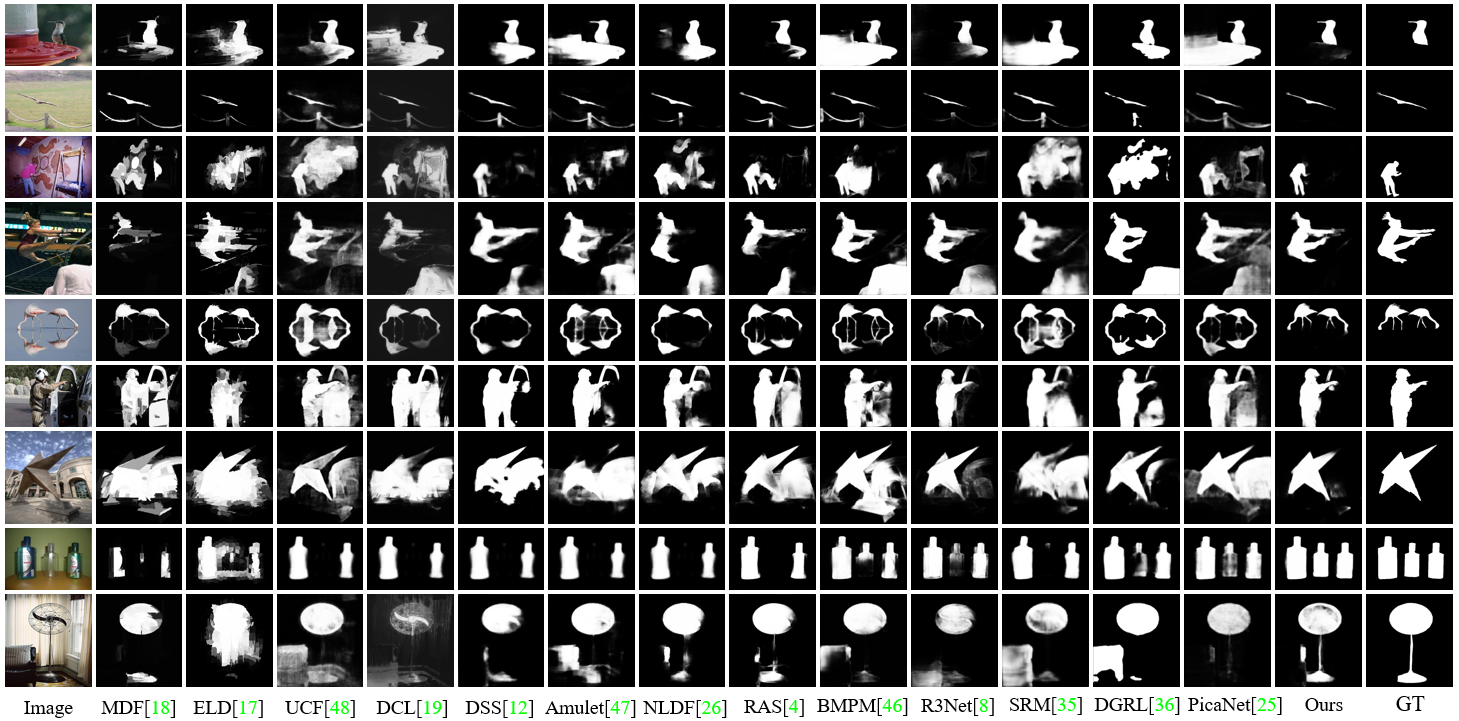}}
%{\includegraphics[width=1.0\textwidth]{pictures/frame1.png}}
\vspace{-4mm}
%\captionsetup{margin=10pt,justification=justified}
\caption{Visual comparison of 13 state-of-the-art deep saliency detection methods.
As can be seen, our model (Ours) produces more coherent and accurate
saliency maps than all other methods, which are the closest to ground truth (GT).}\label{fig:visualMap}
\end{center}
\vspace{-4mm}
\end{figure*}

%%state-of-the-art F-measure and MAE values.
\begin{table*}[pt]
\small
\small
\begin{center}
\resizebox{0.9\textwidth}{!}{
\begin{tabular}{l|c c|c c|c c|c c|c c|c c}
\toprule
\centering
\multirow{2}*{Methods}&\multicolumn{2}{c|}{ECSSD \cite{ECSSDYan}}&\multicolumn{2}{c|}{PASCAL-S \cite{PSACALSLi}}&\multicolumn{2}{c|}{DUTS-test \cite{DUTS}}&\multicolumn{2}{c|}{HKU-IS \cite{MDF:li2015visual}}&\multicolumn{2}{c|}{SOD \cite{SODVida}}&\multicolumn{2}{c}{DUT-OMRON \cite{DUT-OMRYang}}\\
%\cline{2-13}
 & $F_{\beta}$-max & MAE & $F_{\beta}$-max & MAE & $F_{\beta}$-max & MAE & $F_{\beta}$-max & MAE & $F_{\beta}$-max & MAE & $F_{\beta}$-max & MAE \\[1pt]
 \midrule
 \multicolumn{13}{l}{VGG  backbone} \\[1pt]
 \midrule
% \multicolumn{11}{|c|}{ResNet \cite{resnet:He2015Deep} backbone} \\
%\cline{1-11}
MDF {\tiny{CVPR2015}} ~\cite{MDF:li2015visual} & 0.832 & 0.105 & 0.768 & 0.146 & 0.730 & 0.094 & 0.861 & 0.129 & 0.787 & 0.159 & 0.694 & 0.092 \\[1pt]

ELD {\tiny{CVPR2016}} ~\cite{ELD:lee2016deep}& 0.867 & 0.079 & 0.773 & 0.123 & 0.738 & 0.093 & 0.839 & 0.074 & 0.764 & 0.155 & 0.715 & 0.092
\\[1pt]

DS {\tiny{TIP2016}} ~\cite{DS:li2016deepsaliency} & 0.882 & 0.122 & 0.765 & 0.176 & 0.777 & 0.090 & 0.865 & 0.080 & 0.784 & 0.190 & 0.745 & 0.120 \\[1pt]

DCL {\tiny{CVPR2016}} ~\cite{DCL:li2016deep} & 0.890 & 0.088 & 0.805 & 0.125 & 0.782 & 0.088 & 0.885 & 0.072 & 0.823 & 0.141 & 0.739 & 0.097 \\[1pt]

DHS {\tiny{CVPR2016}} ~\cite{DHSNet:liu2016dhsnet} & 0.907 & 0.059 & 0.829 & 0.094 & 0.807 & 0.067 & 0.890 & 0.053 & 0.827 & 0.128 & - & -\\[1pt]

UCF {\tiny{ICCV2017}} ~\cite{UCF:zhang2017learning} & 0.911 & 0.078 & 0.828 & 0.126 & 0.771 & 0.117 & 0.886 & 0.074 & 0.803 & 0.164 & 0.734 & 0.132 \\[1pt]

DSS {\tiny{CVPR2017}} ~\cite{DSS:hou2017deeply} & 0.916 & 0.053 & 0.836 & 0.096 & 0.825 & 0.057 & 0.911 & 0.041 & 0.844 & 0.121 & 0.771 & 0.066 \\[1pt]

NLDF {\tiny{CVPR2017}} ~\cite{NLDF:luo2017non} & 0.905 & 0.063 & 0.831 & 0.099 & 0.812 & 0.066 & 0.902 & 0.048 & 0.841 & 0.124 & 0.753 & 0.080 \\[1pt]

Amulet {\tiny{ICCV2017}} ~\cite{Amulet:zhang2017amulet} & 0.915 & 0.059 & 0.837 & 0.098 & 0.778 & 0.085 & 0.895 & 0.052 & 0.806 & 0.141 & 0.742 & 0.098 \\[1pt]

RAS {\tiny{ECCV2018}} ~\cite{RAS:chen2018eccv} & 0.921 & 0.056 & 0.837 & 0.104 & 0.831 & 0.060 &  0.913 & 0.045 & 0.850 & 0.124 & 0.786 &\textcolor{aa}{\textbf{0.062}} \\[1pt]

BMPM {\tiny{CVPR2018}} ~\cite{BMPM:zhang2018bi} & 0.929 & \textcolor{bb}{\textbf{0.045}} & 0.862 & \textcolor{aa}{\textbf{0.074}} & 0.851 & \textcolor{bb}{\textbf{0.049}} & 0.921 & \textcolor{bb}{\textbf{0.039}} & \textcolor{bb}{\textbf{0.855}} & 0.107 & 0.774 &\textcolor{bb}{\textbf{0.064}} \\[1pt]

PAGR {\tiny{CVPR2018}} ~\cite{PAGR:zhang2018progressive} & 0.927 & 0.061 & 0.856 & 0.093 & \textcolor{bb}{\textbf{0.855}} & 0.056 & 0.918 & 0.048 & - & - & 0.771 & 0.071 \\[1pt]

PicaNet {\tiny{CVPR2018}} ~\cite{PiCANet:liu2018picanet} & \textcolor{bb}{\textbf{0.931}} & 0.047 & \textcolor{bb}{\textbf{0.868}} & \textcolor{bb}{\textbf{0.077}} & 0.851 & 0.054 & \textcolor{bb}{\textbf{0.921}} & 0.042 & 0.853 & \textcolor{bb}{\textbf{0.102}} & \textcolor{bb}{\textbf{0.794}} & 0.068 \\[1pt]

\textbf{SRNet-V} &\textcolor{aa}{\textbf{0.939}} & \textcolor{aa}{\textbf{0.045}} &\textcolor{aa}{\textbf{0.869}} & 0.078
& \textcolor{aa}{\textbf{0.876}} & \textcolor{aa}{\textbf{0.046}}
& \textcolor{aa}{\textbf{0.931}} & \textcolor{aa}{\textbf{0.037}}
& \textcolor{aa}{\textbf{0.859}} & \textcolor{aa}{\textbf{0.082}}
& \textcolor{aa}{\textbf{0.808}} & 0.065 \\[1pt]

\midrule
 \multicolumn{13}{l}{ResNet  backbone} \\[1pt]
\midrule
SRM {\tiny{ICCV2017}} ~\cite{SRM:wang2017stagewise} & 0.917 & 0.054 & 0.847 & 0.085 & 0.827 & 0.059 & 0.906 & 0.046& 0.843 & 0.127 & 0.769 & 0.069 \\[1pt]
%\hline

DGRL {\tiny{CVPR2018}} ~\cite{DGRL:wang2018detect} & 0.922 & \textcolor{bb}{\textbf{0.041}} & 0.854 & \textcolor{bb}{\textbf{0.078}} & 0.829 & 0.056 & 0.910 & \textcolor{bb}{\textbf{0.036}} & 0.845 & \textcolor{bb}{\textbf{0.104}} & 0.774 & 0.062 \\[1pt]

R3Net {\tiny{IJCAI2018}} ~\cite{R3Net:deng18r} & 0.931 & 0.046 & 0.845 & 0.097 & 0.828 & 0.059 & 0.917 & 0.038 & 0.836 & 0.136 & 0.792 & \textcolor{bb}{\textbf{0.061}} \\[1pt]
%\hline
%
PicaNet-R {\tiny{CVPR2018}} ~\cite{PiCANet:liu2018picanet} & \textcolor{bb}{\textbf{0.935}} & 0.047 & \textcolor{aa}{\textbf{0.881}} & 0.087 & \textcolor{bb}{\textbf{0.860}} & \textcolor{bb}{\textbf{0.051}} & \textcolor{bb}{\textbf{0.919}} & 0.043 & \textcolor{bb}{\textbf{0.858}} & 0.109 & \textcolor{bb}{\textbf{0.803}} & 0.065 \\[1pt]
%\hline
%
\textbf{SRNet-R} & \textcolor{aa}{\textbf{0.948}} & \textcolor{aa}{\textbf{0.038}}
& \textcolor{bb}{\textbf{0.877}} & \textcolor{aa}{\textbf{0.074}} & \textcolor{aa}{\textbf{0.888}}
& \textcolor{aa}{\textbf{0.041}} & \textcolor{aa}{\textbf{0.937}} & \textcolor{aa}{\textbf{0.033}}
& \textcolor{aa}{\textbf{0.875}} & \textcolor{aa}{\textbf{0.069}} & \textcolor{aa}{\textbf{0.839}}
& \textcolor{aa}{\textbf{0.056}} \\[1pt]
\bottomrule
\end{tabular}}
\captionsetup{margin=10pt,justification=justified}
\caption{Comparisons of max F-measure ($F_{\beta}$-max) and MAE values.
Results on both VGG \cite{VGG} and ResNet \cite{resnet:He2015Deep} backone are reported,
and the top two results are shown in \textcolor{aa}{\textbf{red}} and \textcolor{bb}{\textbf{blue}} colors, respectively. Best viewed in color.} \label{allCompare}
\end{center}
\vspace{-5mm}
\end{table*}

\section{Experiments}

\subsection{Setup}

\paragraph{Datasets}
We conduct experiments on six widely used saliency detection benchmark datasets,
including ECSSD~\cite{ECSSDYan}, PASCAL-S~\cite{PSACALSLi}, DUT-OMRON~\cite{DUT-OMRYang},
HKU-IS~\cite{MDF:li2015visual}, SOD~\cite{SODVida} and DUTS-test~\cite{DUTS}.
These datasets provide 1000, 850, 5,168, 4,447, 300 and 5,019 natural images of complex contents respectively,
with manually labeled pixel-wise saliency ground-truth.

\vspace{-4mm}
\paragraph{Implementation} We train the proposed SRNet on the training split of the DUTS dataset~\cite{DUTS}.
All the training images are resized to $320 \times 320$.
We use random rotation and and horizontal flipping to augment the training data.
We train SRNet with
learning rate 0.01, weight decay 0.0005, and momentum 0.9.
We adopt the pre-trained Resnet101~\cite{resnet:He2015Deep} and VGG16~\cite{VGG}
as initialized backbones,
and denote corresponding models as SRNet-R and SRNet-V respectively.
The middle output channels of the saliency reasoning module are $\{64,48,32\}$, respectively.
Correspondingly,
for the first branch in the saliency reasoning module,
the dilation rates for the $3\times3$ depth-wise convolution are $\{2,2,1\}$.
Meanwhile,
$\{1,1,2\}$ and $\{3,7,3\}$ SR-unit are adopted for training SRNet-R and SRNet-V, respectively.
\vspace{-4mm}

\paragraph{Evaluation metrics}
Following recent studies~\cite{Amulet:zhang2017amulet,PAGR:zhang2018progressive,DSS:hou2017deeply,
DHSNet:liu2016dhsnet,DGRL:wang2018detect,UCF:zhang2017learning,NLDF:luo2017non,RAS:chen2018eccv,R3Net:deng18r},
we adopt the widely used precision-recall (PR) curves, F-measure ($F_{\beta}$-max), and
mean absolute error (MAE) as  evaluation metrics.
Detailed descriptions of these metrics can be seen in~\cite{DSS:hou2017deeply}\cite{Borjisuvery}.

%%%state-of-the-art PR curves.
\begin{figure*}[!pt]
\begin{center}
\subfigure
{\includegraphics[width=0.31\textwidth,height=0.24\textwidth]{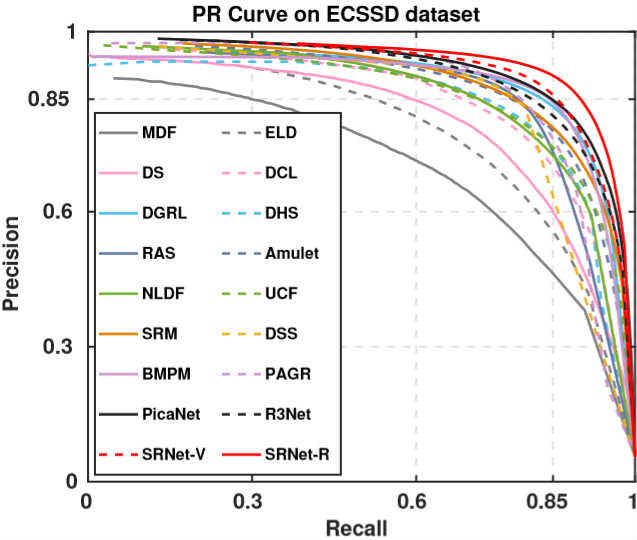}}
\subfigure
{\includegraphics[width=0.31\textwidth,height=0.24\textwidth]{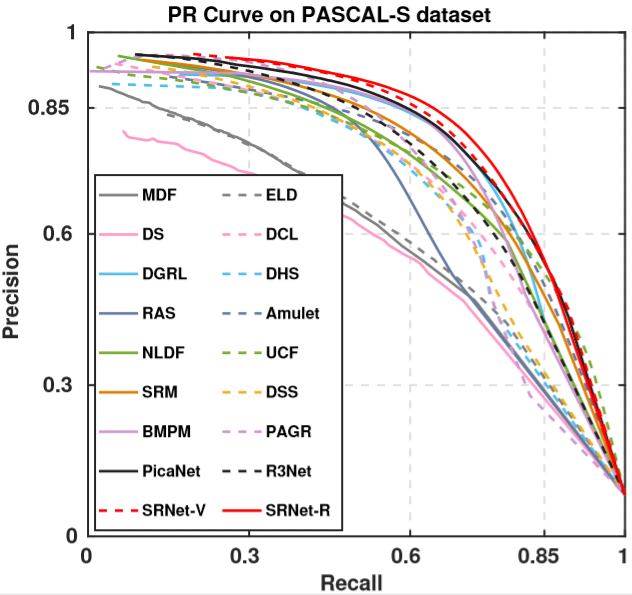}}
\subfigure
{\includegraphics[width=0.31\textwidth,height=0.24\textwidth]{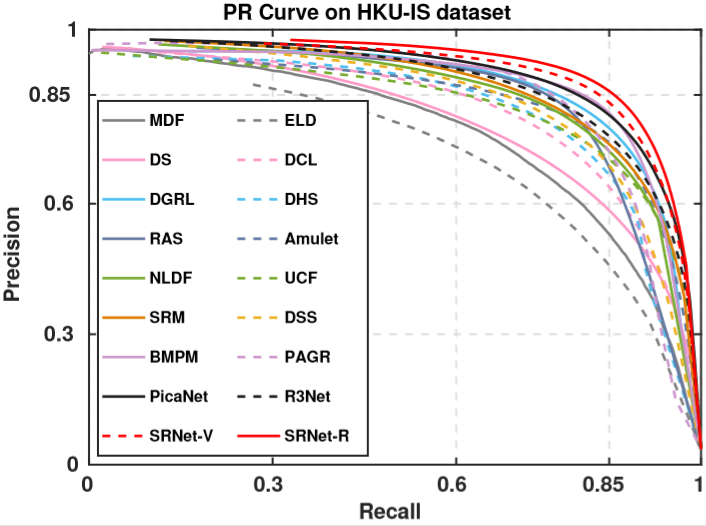}}
\subfigure
{\includegraphics[width=0.31\textwidth,height=0.24\textwidth]{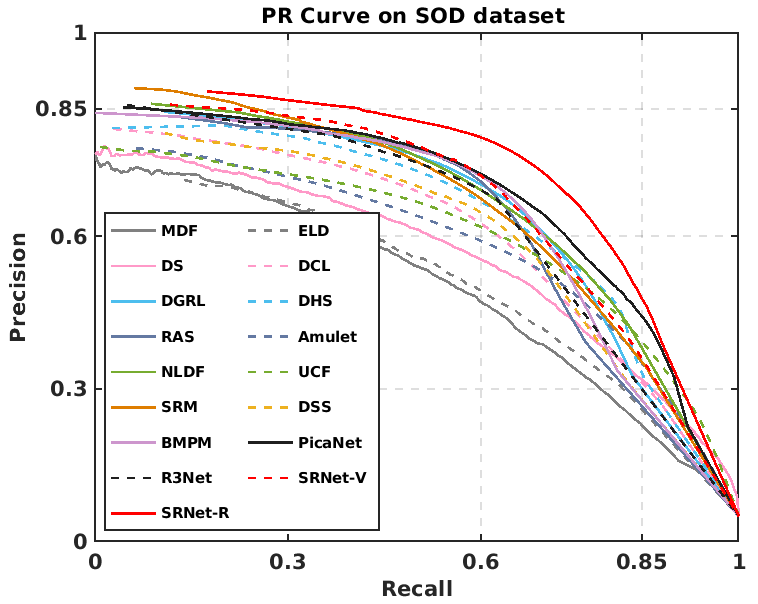}}
\subfigure
{\includegraphics[width=0.31\textwidth,height=0.24\textwidth]{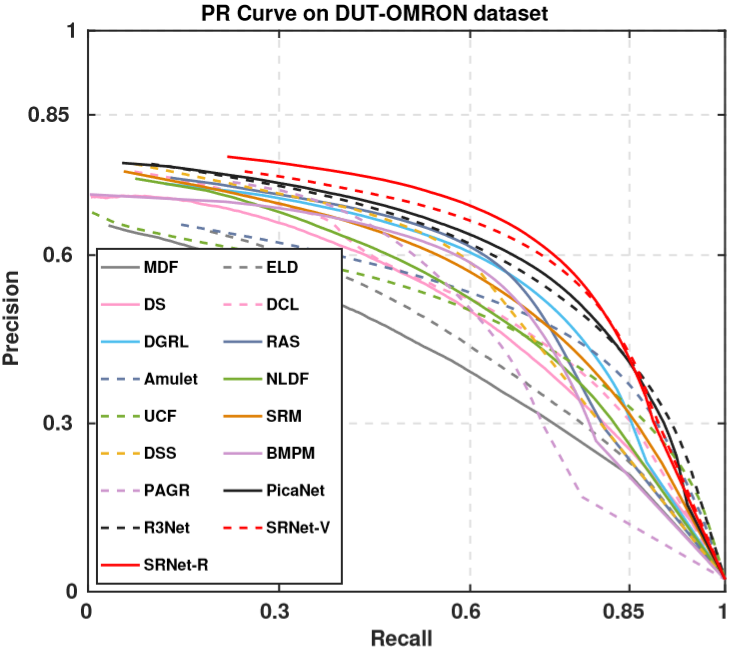}}
\subfigure
{\includegraphics[width=0.31\textwidth,height=0.24\textwidth]{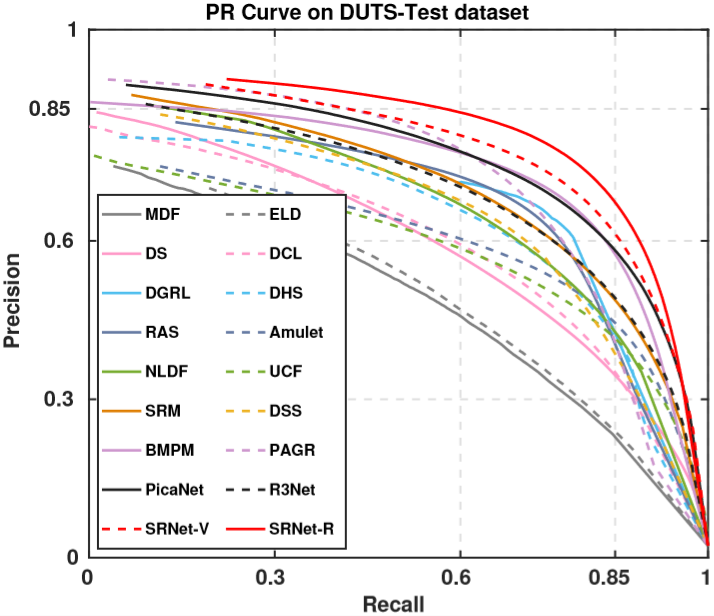}}
\vspace{-4mm}
\captionsetup{margin=10pt,justification=justified}
\caption{Quantitative results of PR curves for the instantiated model and other state-of-the-art models.
The designed models of SRNet-V and SRNet-R take VGG16 \cite{VGG} and ResNet101 \cite{resnet:He2015Deep} as backbone respectively,
they consistently outperform other models across all the testing datasets.
}\label{fig:allPR}
\end{center}
\vspace{-4mm}
\end{figure*}

\subsection{Comparison with state-of-the-arts}

We compare our SRNet with
recent 16 deep CNN-based saliency models (given in Table~\ref{allCompare} for brevity).
For fairness, we adopt the comparison results
provided by~\cite{Feng:sal_eval_toolbox} for all baselines.
The results of DHS~\cite{MDF:li2015visual} on DUT-OMRON \cite{DUT-OMRYang} are
not reported because it uses a part of DUT-OMRON for training.
Similarly, we do not report PAGR~\cite{PAGR:zhang2018progressive} results on SOD \cite{SODVida}.

\vspace{-4mm}
\paragraph{Visual comparison}
Fig.~\ref{fig:visualMap} gives visual comparisons of SRNet (Ours)
with state-of-the-arts.
One can see that SRNet predicts salient object maps closest to the ground-truth in
various challenging scenarios,
such as images with complex backgrounds and foregrounds (row 3, 4, 6 and 7),
objects having similar appearance with background (row 1, 2, and 9), and
multiple instances of the same object (row 5). Our model can well segment the entire objects (row 1, 4 and 8) with fine saliency details (row 5, 7 and 9),
demonstrating the effectiveness and benefits  of the proposed deep reasoning module for salient object detection.

\vspace{-4mm}
\paragraph{F-measure and MAE}
Table \ref{allCompare} reports $F_{\beta}$-max and MAE scores of SRNet-R and SRNet-V models
on six datasets, and comparisons with the baselines.
The SRNet significantly outperforms others on most of the datasets.
Specifically,
comparing the models using VGG backbone in the $F_{\beta}$-max scores,
SRNet-V outperforms the best baseline by  0.8\%, 1.9\%, 1.0\%, 1.4\% on the
ECSSD, DUTS-test, HKU-IS, and DUT-OMRON datasets, respectively.
Besides,
as for the MAE metric,
SRNet-V still ranks the first even on the challenging datasets HKUIS, SOD, and DUT-OMRON.
When changing the backbone to ResNet~\cite{resnet:He2015Deep},
we observe more significant performance improvement brought by SRNet-R over other ResNet based models. In particular, the $F_{\beta}$-max scores of SRNet-R are 1.3\%, 2.8\%, 1.8\%, 1.7\%, and 3.6\%
higher than the second best baselines on the ECSSD, DUTS-test, HKU-IS, SOD
and DUT-OMRON datasets, respectively.
Also, the MAE scores of SRNet-R are 0.4\%, 1.0\%, 3.5\%, 0.5\% lower than the
second best method on the PASCALS, DUTS-test, SOD and DUT-OMRON datasets, respectively,
illustrating the effectiveness of the SRNet.

%%Ablation tabel
\begin{table*}[!pt]
\small
\begin{center}
\resizebox{0.9\textwidth}{!}{
\begin{tabular}{l|c c|c c|c c|c c|c c|c c}
\toprule
\centering
\multirow{2}*{Module}&\multicolumn{2}{c|}{ECSSD \cite{ECSSDYan}}&\multicolumn{2}{c|}{PASCAL-S \cite{PSACALSLi}}&\multicolumn{2}{c|}{DUTS-test \cite{DUTS}}&\multicolumn{2}{c|}{HKU-IS \cite{MDF:li2015visual}}&\multicolumn{2}{c|}{SOD \cite{SODVida}}&\multicolumn{2}{c}{DUT-OMRON \cite{DUT-OMRYang}}\\
%\cline{2-13}
 & $F_{\beta}$-max & MAE & $F_{\beta}$-max & MAE & $F_{\beta}$-max & MAE & $F_{\beta}$-max & MAE & $F_{\beta}$-max & MAE & $F_{\beta}$-max & MAE \\[1pt]
 \midrule
 \multicolumn{13}{l}{VGG  backbone} \\[1pt]
 \midrule
% \multicolumn{11}{|c|}{ResNet \cite{resnet:He2015Deep} backbone} \\
%\cline{1-11}
BPS & 0.899 & 0.094 & 0.845 & 0.108 & 0.816 & 0.097 & 0.889 & 0.087 & 0.812 & 0.122 & 0.746 & 0.123 \\[1pt]

HFS &  0.922 &  0.056 &  0.865 &  0.081 &  0.854 &  0.056 &  0.915 &  0.046 &  0.842 &  0.091 &  0.776 &  0.079 \\[1pt]

BFR & \textcolor{bb}{\textbf{0.938}} & \textcolor{bb}{\textbf{0.045}} & \textcolor{bb}{\textbf{0.868}} & \textcolor{bb}{\textbf{0.078}} & \textcolor{bb}{\textbf{0.869}} & \textcolor{bb}{\textbf{0.047}} & \textcolor{bb}{\textbf{0.929}} & \textcolor{bb}{\textbf{0.038}} & \textcolor{bb}{\textbf{0.851}} & \textcolor{bb}{\textbf{0.084}} & \textcolor{bb}{\textbf{0.802}} & \textcolor{bb}{\textbf{0.067}} \\[1pt]

\textbf{SRNet-V} & \textcolor{aa}{\textbf{0.939}} & \textcolor{aa}{\textbf{0.045}} &\textcolor{aa}{\textbf{0.869}} & \textcolor{aa}{\textbf{0.078}}
& \textcolor{aa}{\textbf{0.876}} & \textcolor{aa}{\textbf{0.046}}
& \textcolor{aa}{\textbf{0.931}} & \textcolor{aa}{\textbf{0.037}}
& \textcolor{aa}{\textbf{0.859}} & \textcolor{aa}{\textbf{0.082}}
& \textcolor{aa}{\textbf{0.808}} & \textcolor{aa}{\textbf{0.065}} \\[1pt]

\midrule
 \multicolumn{13}{l}{ResNet  backbone} \\[1pt]
\midrule
BPS & 0.903 & 0.063 & 0.846 & 0.087 & 0.813 & 0.063 & 0.882 & 0.060 & 0.812 & 0.093 & 0.773 & 0.076 \\[1pt]

HFS &  0.937 & 0.049 & 0.873
& \textcolor{aa}{\textbf{0.070}} & 0.877 & 0.043 & 0.924 & 0.037 & 0.849 & 0.081 & 0.820 & 0.063\\[1pt]
BFR & \textcolor{bb}{\textbf{0.946}} & \textcolor{bb}{\textbf{0.039}} & \textcolor{aa}{\textbf{0.879}} & \textcolor{bb}{\textbf{0.073}} & \textcolor{bb}{\textbf{0.886}} & \textcolor{bb}{\textbf{0.042}} & \textcolor{bb}{\textbf{0.937}} & \textcolor{bb}{\textbf{0.034}} & \textcolor{bb}{\textbf{0.867}} & \textcolor{bb}{\textbf{0.071}} & \textcolor{bb}{\textbf{0.832}} & \textcolor{bb}{\textbf{0.058}} \\[1pt]

\textbf{SRNet-R} & \textcolor{aa}{\textbf{0.948}} & \textcolor{aa}{\textbf{0.038}}
& \textcolor{bb}{\textbf{0.877}} & 0.074 & \textcolor{aa}{\textbf{0.888}}
& \textcolor{aa}{\textbf{0.041}} & \textcolor{aa}{\textbf{0.937}} & \textcolor{aa}{\textbf{0.033}}
& \textcolor{aa}{\textbf{0.875}} & \textcolor{aa}{\textbf{0.069}} & \textcolor{aa}{\textbf{0.839}}
& \textcolor{aa}{\textbf{0.056}}\\[1pt]
\bottomrule
\end{tabular}}
\small\captionsetup{margin=20pt,justification=justified}
\caption{Evaluation results of $F_{\beta}$-max and MAE with different modules on 6 datasets for ablation studies. We report these results on both VGG \cite{VGG} and ResNet \cite{resnet:He2015Deep} backbone.
The top two results are highlighted in \textcolor{aa}{\textbf{red}} and \textcolor{bb}{\textbf{blue}} colors. Best viewed in color.}\label{ablationCompare}
\vspace{-4mm}
\end{center}
\vspace{-2mm}
\end{table*}
\vspace{-4mm}
\paragraph{PR curves}
Fig.~\ref{fig:allPR} plots PR curves of SRNet with two different backbones, SRNet-R \& SRNet-V, as well as
 baseline models on six datasets.
It can be clearly seen that our model consistently outperforms all baseline  models across all datasets, justifying the effectiveness of developing a deeper saliency reasoning and inference module.
Besides,
compared with SRNet-V,
 SRNet-R gives significantly higher PR-curves,
especially on the challenging DUTS-Test, SOD, and DUT-OMRON. This demonstrates
 the superior performance of our model when performing saliency reasoning from
more comprehensive features (offered by a deeper backbone).
%%%layers curves.
\begin{figure*}[!pt]
\begin{center}
\subfigure
{\includegraphics[width=0.24\textwidth,height=0.2\textwidth]{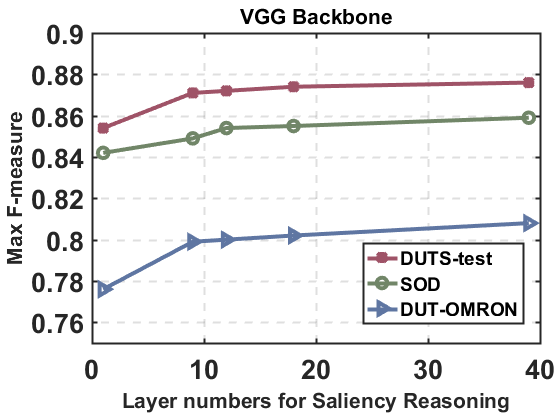}}
\subfigure
{\includegraphics[width=0.24\textwidth,height=0.2\textwidth]{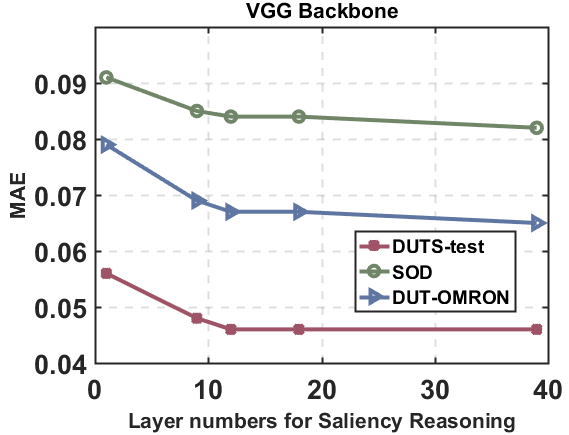}}
\subfigure
{\includegraphics[width=0.24\textwidth,height=0.2\textwidth]{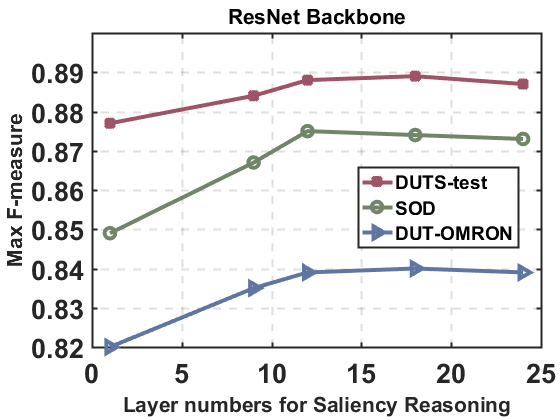}}
\subfigure
{\includegraphics[width=0.24\textwidth,height=0.2\textwidth]{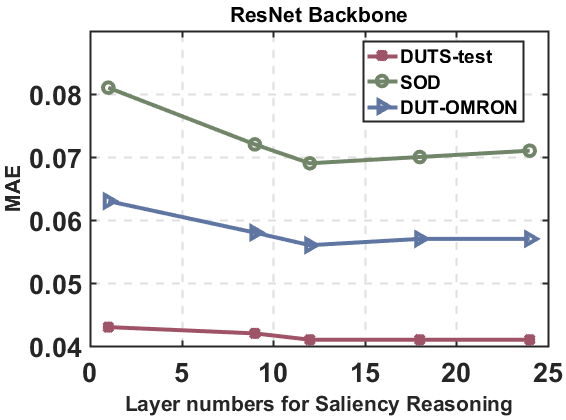}}
\vspace{-4mm}
\captionsetup{margin=10pt,justification=justified}
\caption{Quantitative results of F-measure and MAE with different number of layers by taking VGG16 \cite{VGG} and ResNet101 \cite{resnet:He2015Deep} as backbone respectively.
}\label{fig:tends}
\end{center}
\vspace{-8mm}
\end{figure*}

%%% backbone visual map.
\begin{figure}[!pt]
\begin{center}
\subfigure
{\includegraphics[width=0.44\textwidth,height=0.23\textwidth]{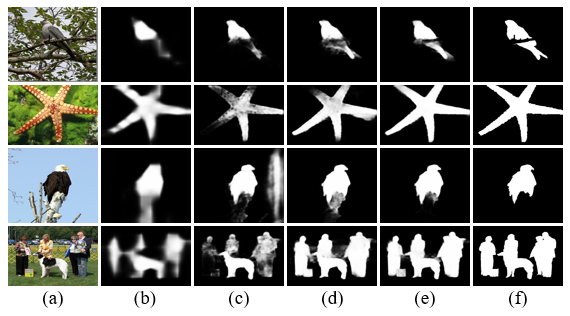}}
\vspace{-4mm}
\captionsetup{margin=5pt,justification=justified}
\caption{Visual comparisons of our model against baseline ResNet~\cite{resnet:He2015Deep}.
(a) Input images. (b) (c) (d) (e) are saliency maps produced by BPS, HFS, BFR and SRNet-R, respectively.
(f) The ground truth.}\label{fig:BackboneMap}
\end{center}
\vspace{-6mm}
\end{figure}

\subsection{Ablation analysis}
\paragraph{Saliency reasoning module}
%We take ResNet \cite{resnet:He2015Deep} and VGG \cite{VGG} as the backbone,
We analyze contributions of deep saliency reasoning with several ablation modules.
For simplicity,
we denote BPS as saliency module with backbone only,
HFS as that with hierarchical feature fusion,
and BFR as a SRNet variant without dilated  convolution.

The $F_{\beta}$-max and MAE scores of different variants are reported in Table~\ref{ablationCompare}.
One can clearly find BFR uniformly outperforms HFS and BPS on most datasets,
implying benefits of a strong saliency reasoning module for saliency detection.
Particularly,
with VGG backbone,
$F_{\beta}$-max scores of BFR are 1.6\%, 1.5\%, 1.4\%, 0.9\% and 2.6\% higher than HFS,
and 3.9\%, 5.3\%, 4.0\%, 3.9\% and 5.6\% higher than BPS,
on ECSSD, DUTS-test, HKU-IS, and DUT-OMRON datasets, respectively.
Similar observations can be made for the MAE metric.
Moreover,
with ResNet backbone,
BFR significantly outperforms BPS with a large margin w.r.t\ both evaluated metrics.
These results convincingly show effectiveness of saliency reasoning.
In Table~\ref{ablationCompare},
one can also find by adding dilated saliency reasoning,
SRNet performs better than BFR, especially on the challenging SOD and DUT-OMRON.
This speaks well for the effectiveness of the varying dilation in saliency reasoning module.

We also display some visual maps in Fig.~\ref{fig:BackboneMap} from variants of SRNet.
The saliency maps generated from BPS and HFS are incomplete and some details of the salient objects are missing.
But our full SRNet model highlights the salient objects completely even for challenging samples.

\vspace{-4mm}
\paragraph{Depth of saliency reasoning module}
We further investigate effects of varying the depth of the reasoning module upon detection performance.
Here,
we train SRNet-R with $1, 9, 12, 18, 24$ depth-wise convolutional layers, and SRNet-V with
$1, 9, 12, 18, 39$ depth-wise convolutional layers.
Fig.~\ref{fig:tends} gives the F-measure and MAE scores over the DUTS-test, SOD and DUT-OMRON datasets.
It can be seen the performance increases with he increasing depth of the saliency reasoning module.
This supports our findings in Sec.~\ref{intro} that a reasoning module with larger capacity would further boost saliency detection performance.
%% layers visual maps
\begin{figure}[!pt]
\begin{center}
\subfigure
{\includegraphics[width=0.42\textwidth,height=0.24\textwidth]{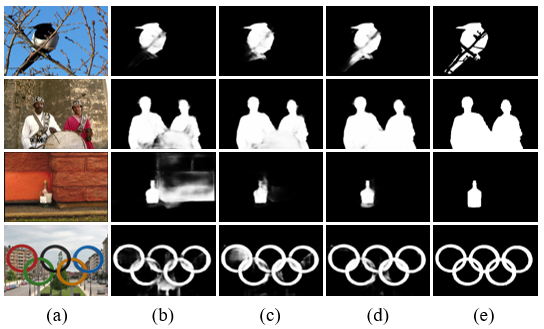}}
\vspace{-4mm}
\captionsetup{margin=10pt,justification=justified}
\caption{Visual comparison of saliency detection with different numbers of convolutional layers for performing saliency reasoning. (a), (e) are input images and their ground truth maps.
(b) (c) (d) show saliency maps of SRNet-R implemented with $3, 9, 12$ dilated depth-wise convolutional layers in saliency reasoning module, respectively.}
\label{fig:LayerMap}
\end{center}
\vspace{-7mm}
\end{figure}

Fig.~\ref{fig:LayerMap} shows the qualitative results.
We find that more precise salient objects can be detected by progressively adding
the convolutional layers in saliency reasoning module. It detects multiple objects,
suppresses non-salient regions, and produces complete salient objects effectively.

\vspace{-4mm}
\paragraph{Computational efficiency of deep reasoning}
We also experiment on a single NVIDIA TITAN X GPU to test the computational efficiency of SRNet.
For an image with size $480\times320$,
Table \ref{timecost} reports its running time of different ablation modules.
Obviously,
although with a deep saliency reasoning module,
SRNet has an equivalent computational speed as BFR and HFS.
This is mainly because of the depth-wise and $1\times1$ group convolutions in SRNet,
which has less parameters but with high saliency reasoning accuracy.

%%time analyses
\vspace{-2mm}
\begin{table}[!ht]
\small
\begin{center}
\begin{tabular}{c c|c c}
\toprule[1.0pt]
\multicolumn{2}{c|}{VGG  backbone}&\multicolumn{2}{c}{ResNet  backbone} \\[1pt]
\midrule
Module & Time (s) & Module & Time (s) \\
\midrule
BPS & 0.265 & BPS & 0.260 \\
HFS & 0.267 & HFS & 0.272 \\
BFR & 0.271 & BFR & 0.274 \\
\textbf{SRNet-V} & \textbf{0.272} & \textbf{SRNet-R} & \textbf{0.275} \\
\bottomrule[1.0pt]
\end{tabular}
\end{center}
\vspace{-4mm}
\captionsetup{margin=20pt,justification=justified}
\caption{Run time analysis of SRNet and the ablation modules.
Results of SRNet are shown as \textbf{bold} fonts.}\label{timecost}
\vspace{-4mm}
\end{table}

\vspace{-2mm}
\section{Conclusion}

In this paper,
we show the saliency inference component that predicts salient regions from the fused features is critical for accurate saliency detection.
A deep and light-weight saliency reasoning module that adopts multi-dilated depth-wise convolutions  is proposed, which directly performs reasoning about salient objects from multi-scale features fast. Comprehensive experiments demonstrate that our method
gives superior salient object detection performance with lower computation time, outperforming state-of-the-art approaches.

\bibliographystyle{ieee}
\bibliography{egbib}

\begin{thebibliography}{10}\itemsep=-1pt

\bibitem{FT}
R.~Achanta, S.~Hemami, F.~Estrada, and S.~S{\"u}sstrunk.
\newblock Frequency-tuned salient region detection.
\newblock In {\em CVPR}, pages 1597--1604, 2009.

\bibitem{RSD}
M.~Amirul~Islam, M.~Kalash, and N.~D. Bruce.
\newblock Revisiting salient object detection: Simultaneous detection, ranking,
  and subitizing of multiple salient objects.
\newblock In {\em CVPR}, pages 7142--7150, 2018.

\bibitem{Borjisuvery}
A.~Borji, D.~N. Sihite, and L.~Itti.
\newblock Salient object detection: a benchmark.
\newblock {\em TIP}, 24(12):5706--5722, 2015.

\bibitem{RAS:chen2018eccv}
S.~Chen, X.~Tan, B.~Wang, and X.~Hu.
\newblock Reverse attention for salient object detection.
\newblock In {\em ECCV}, pages 236--252, 2018.

\bibitem{GlorNet}
Y.~Chen, M.~Rohrbach, Z.~Yan, S.~Yan, J.~Feng, and Y.~Kalantidis.
\newblock Graph-based global reasoning networks.
\newblock {\em arXiv preprint arXiv:1811.12814}, 2018.

\bibitem{GC}
M.-M. Cheng, N.~J. Mitra, X.~Huang, P.~H.~S. Torr, and S.-M. Hu.
\newblock Global contrast based salient region detection.
\newblock {\em TPAMI}, 37(3):569--582, 2015.

\bibitem{Xception}
F.~Chollet.
\newblock Xception: Deep learning with depthwise separable convolutions.
\newblock In {\em CVPR}, pages 1251--1258, 2017.

\bibitem{R3Net:deng18r}
Z.~Deng, X.~Hu, L.~Zhu, X.~Xu, J.~Qin, G.~Han, and P.-A. Heng.
\newblock R$^{3}${N}et: Recurrent residual refinement network for saliency
  detection.
\newblock In {\em IJCAI}, pages 684--690, 2018.

\bibitem{captionII}
F.~Hao, S.~Gupta, F.~Iandola, R.~Srivastava, D.~Li, P.~Dollár, J.~Gao, X.~He,
  M.~Mitchell, and J.~C. Platt.
\newblock From captions to visual concepts and back.
\newblock In {\em CVPR}, pages 1473--1482, 2015.

\bibitem{resnet:He2015Deep}
K.~He, X.~Zhang, S.~Ren, and J.~Sun.
\newblock Deep residual learning for image recognition.
\newblock In {\em CVPR}, pages 770--778, 2015.

\bibitem{tracking}
S.~Hong, T.~You, S.~Kwak, and B.~Han.
\newblock Online tracking by learning discriminative saliency map with
  convolutional neural network.
\newblock In {\em ICML}, 2015.

\bibitem{DSS:hou2017deeply}
Q.~Hou, M.-M. Cheng, X.~Hu, A.~Borji, Z.~Tu, and P.~Torr.
\newblock Deeply supervised salient object detection with short connections.
\newblock In {\em CVPR}, pages 5300--5309, 2017.

\bibitem{MobileNets}
A.~G. Howard, M.~Zhu, C.~Bo, D.~Kalenichenko, W.~Wang, T.~Weyand, M.~Andreetto,
  and H.~Adam.
\newblock Mobilenets: Efficient convolutional neural networks for mobile vision
  applications.
\newblock {\em arXiv preprint arXiv:1704.04861}, 2017.

\bibitem{SqueezeNet}
F.~N. Iandola, S.~Han, M.~W. Moskewicz, K.~Ashraf, W.~J. Dally, and K.~Keutzer.
\newblock Squeezenet: Alexnet-level accuracy with 50x fewer parameters and $<$
  0.5 mb model size.
\newblock {\em arXiv preprint arXiv:1602.07360}, 2016.

\bibitem{DRFI}
H.~Jiang, J.~Wang, Z.~Yuan, Y.~Wu, N.~Zheng, and S.~Li.
\newblock Salient object detection: A discriminative regional feature
  integration approach.
\newblock In {\em CVPR}, pages 2083--2090, 2013.

\bibitem{AlexNet}
A.~Krizhevsky, I.~Sutskever, and G.~E. Hinton.
\newblock Imagenet classification with deep convolutional neural networks.
\newblock In {\em Advances in neural information processing systems}, pages
  1097--1105, 2012.

\bibitem{ELD:lee2016deep}
G.~Lee, Y.-W. Tai, and J.~Kim.
\newblock Deep saliency with encoded low level distance map and high level
  features.
\newblock In {\em CVPR}, pages 660--668, 2016.

\bibitem{MDF:li2015visual}
G.~Li and Y.~Yu.
\newblock Visual saliency based on multiscale deep features.
\newblock In {\em CVPR}, pages 5455--5463, 2015.

\bibitem{DCL:li2016deep}
G.~Li and Y.~Yu.
\newblock Deep contrast learning for salient object detection.
\newblock In {\em CVPR}, pages 478--487, 2016.

\bibitem{DSR}
X.~Li, H.~Lu, L.~Zhang, R.~Xiang, and M.~H. Yang.
\newblock Saliency detection via dense and sparse reconstruction.
\newblock In {\em ICCV}, 2013.

\bibitem{DS:li2016deepsaliency}
X.~Li, L.~Zhao, L.~Wei, M.-H. Yang, F.~Wu, Y.~Zhuang, H.~Ling, and J.~Wang.
\newblock Deepsaliency: Multi-task deep neural network model for salient object
  detection.
\newblock {\em TIP}, 25(8):3919--3930, 2016.

\bibitem{PSACALSLi}
Y.~Li, X.~Hou, C.~Koch, J.~M. Rehg, and A.~L. Yuille.
\newblock The secrets of salient object segmentation.
\newblock In {\em CVPR}, pages 280--287, 2014.

\bibitem{lin2015bilinear}
T.-Y. Lin, A.~RoyChowdhury, and S.~Maji.
\newblock Bilinear cnn models for fine-grained visual recognition.
\newblock In {\em ICCV}, pages 1449--1457, 2015.

\bibitem{DHSNet:liu2016dhsnet}
N.~Liu and J.~Han.
\newblock Dhsnet: Deep hierarchical saliency network for salient object
  detection.
\newblock In {\em CVPR}, pages 678--686, 2016.

\bibitem{PiCANet:liu2018picanet}
N.~Liu, J.~Han, and M.-H. Yang.
\newblock Picanet: Learning pixel-wise contextual attention for saliency
  detection.
\newblock In {\em CVPR}, pages 3089--3098, 2018.

\bibitem{NLDF:luo2017non}
Z.~Luo, A.~K. Mishra, A.~Achkar, J.~A. Eichel, S.~Li, and P.-M. Jodoin.
\newblock Non-local deep features for salient object detection.
\newblock In {\em CVPR}, pages 6593--6601, 2017.

\bibitem{ShuffleNetV2}
N.~Ma, X.~Zhang, H.-T. Zheng, and J.~Sun.
\newblock Shufflenet v2: Practical guidelines for efficient cnn architecture
  design.
\newblock pages 116--131, 2018.

\bibitem{Feng:sal_eval_toolbox}
{Mengyang Feng}.
\newblock Evaluation toolbox for salient object detection.
\newblock \url{https://github.com/ArcherFMY/sal_eval_toolbox}, 2018.

\bibitem{SODVida}
V.~Movahedi and J.~H. Elder.
\newblock Design and perceptual validation of performance measures for salient
  object segmentation.
\newblock In {\em CVPRW}, pages 49--56, 2010.

\bibitem{MobileNetV2}
M.~Sandler, A.~Howard, M.~Zhu, A.~Zhmoginov, and L.~C. Chen.
\newblock Mobilenetv2: Inverted residuals and linear bottlenecks.
\newblock pages 4510--4520, 2018.

\bibitem{VGG}
K.~Simonyan and A.~Zisserman.
\newblock Very deep convolutional networks for large-scale image recognition.
\newblock {\em CoRR}, abs/1409.1556, 2018.

\bibitem{LEGS:wang2015deep}
L.~Wang, H.~Lu, X.~Ruan, and M.-H. Yang.
\newblock Deep networks for saliency detection via local estimation and global
  search.
\newblock In {\em CVPR}, pages 3183--3192, 2015.

\bibitem{DUTS}
L.~Wang, H.~Lu, Y.~Wang, M.~Feng, D.~Wang, B.~Yin, and X.~Ruan.
\newblock Learning to detect salient objects with image-level supervision.
\newblock In {\em CVPR}, pages 136--145, 2017.

\bibitem{RFCN:wang2016saliency}
L.~Wang, L.~Wang, H.~Lu, P.~Zhang, and X.~Ruan.
\newblock Saliency detection with recurrent fully convolutional networks.
\newblock In {\em ECCV}, pages 825--841, 2016.

\bibitem{SRM:wang2017stagewise}
T.~Wang, A.~Borji, L.~Zhang, P.~Zhang, and H.~Lu.
\newblock A stagewise refinement model for detecting salient objects in images.
\newblock In {\em ICCV}, pages 4019--4028, 2017.

\bibitem{DGRL:wang2018detect}
T.~Wang, L.~Zhang, S.~Wang, H.~Lu, G.~Yang, X.~Ruan, and A.~Borji.
\newblock Detect globally, refine locally: A novel approach to saliency
  detection.
\newblock In {\em CVPR}, pages 3127--3135, 2018.

\bibitem{SDF}
W.~Wang, J.~Shen, X.~Dong, and B.~Ali.
\newblock Salient object detection driven by fixation prediction.
\newblock In {\em CVPR}, pages 1711--1720, 2018.

\bibitem{wang2018videos}
X.~Wang and A.~Gupta.
\newblock Videos as space-time region graphs.
\newblock {\em arXiv preprint arXiv:1806.01810}, 2018.

\bibitem{GS}
Y.~Wei, W.~Fang, W.~Zhu, and S.~Jian.
\newblock Geodesic saliency using background priors.
\newblock In {\em ECCV}, pages 29--42, 2012.

\bibitem{Wei2015STC}
Y.~Wei, X.~Liang, Y.~Chen, X.~Shen, M.~M. Cheng, J.~Feng, Y.~Zhao, and S.~Yan.
\newblock Stc: A simple to complex framework for weakly-supervised semantic
  segmentation.
\newblock {\em TPAMI}, 39(11):2314--2320, 2015.

\bibitem{resnext:Xie2016}
S.~Xie, R.~Girshick, P.~Doll{\'a}r, Z.~Tu, and K.~He.
\newblock Aggregated residual transformations for deep neural networks.
\newblock In {\em CVPR}, pages 1492--1500, 2017.

\bibitem{HED}
S.~Xie and Z.~Tu.
\newblock Holistically-nested edge detection.
\newblock In {\em ICCV}, pages 1395--1403, 2015.

\bibitem{captionI}
K.~Xu, J.~Ba, R.~Kiros, K.~Cho, A.~Courville, R.~Salakhudinov, R.~Zemel, and
  Y.~Bengio.
\newblock Show, attend and tell: Neural image caption generation with visual
  attention.
\newblock In {\em ICML}, pages 2048--2057, 2015.

\bibitem{ECSSDYan}
Q.~Yan, L.~Xu, J.~Shi, and J.~Jia.
\newblock Hierarchical saliency detection.
\newblock In {\em CVPR}, pages 1155--1162, 2013.

\bibitem{DUT-OMRYang}
C.~Yang, L.~Zhang, H.~Lu, X.~Ruan, and M.-H. Yang.
\newblock Saliency detection via graph-based manifold ranking.
\newblock In {\em CVPR}, pages 3166--3173, 2013.

\bibitem{BMPM:zhang2018bi}
L.~Zhang, J.~Dai, H.~Lu, Y.~He, and G.~Wang.
\newblock A bi-directional message passing model for salient object detection.
\newblock In {\em CVPR}, pages 1741--1750, 2018.

\bibitem{Amulet:zhang2017amulet}
P.~Zhang, D.~Wang, H.~Lu, H.~Wang, and X.~Ruan.
\newblock Amulet: Aggregating multi-level convolutional features for salient
  object detection.
\newblock In {\em ICCV}, pages 202--211, 2017.

\bibitem{UCF:zhang2017learning}
P.~Zhang, D.~Wang, H.~Lu, H.~Wang, and B.~Yin.
\newblock Learning uncertain convolutional features for accurate saliency
  detection.
\newblock In {\em ICCV}, pages 212--221, 2017.

\bibitem{PAGR:zhang2018progressive}
X.~Zhang, T.~Wang, J.~Qi, H.~Lu, and G.~Wang.
\newblock Progressive attention guided recurrent network for salient object
  detection.
\newblock In {\em CVPR}, pages 714--722, 2018.

\bibitem{ShuffleNet}
X.~Zhang, X.~Zhou, M.~Lin, and J.~Sun.
\newblock Shufflenet: An extremely efficient convolutional neural network for
  mobile devices.
\newblock In {\em CVPR}, pages 6848--6856, 2018.

\bibitem{MCDL:zhao2015saliency}
R.~Zhao, W.~Ouyang, H.~Li, and X.~Wang.
\newblock Saliency detection by multi-context deep learning.
\newblock In {\em CVPR}, pages 1265--1274, 2015.

\bibitem{understanding}
J.~Y. Zhu, J.~Wu, Y.~Wei, E.~Chang, and Z.~Tu.
\newblock Unsupervised object class discovery via saliency-guided multiple
  class learning.
\newblock {\em TPAMI}, 37(4):862--874, 2014.

\bibitem{RBD}
W.~Zhu, S.~Liang, Y.~Wei, and J.~Sun.
\newblock Saliency optimization from robust background detection.
\newblock In {\em CVPR}, pages 2814--2821, 2014.

\end{thebibliography}
\end{document}